\documentclass[letterpaper, 10 pt, conference]{ieeeconf}  

\IEEEoverridecommandlockouts                              

\usepackage[font=small,labelfont=bf]{caption}
\usepackage[pdftex]{graphicx}
\usepackage{graphics} 
\usepackage{float} 
\usepackage{hyperref} 
\hypersetup{
  colorlinks   = true, 
  urlcolor     = blue, 
  linkcolor    = blue, 
}
\usepackage{epsfig} 
\usepackage{amsmath} 
\usepackage{amssymb}  
\usepackage{enumerate}
\usepackage{amsfonts}
\usepackage{subfig}
\usepackage{epstopdf}
\usepackage{booktabs}
\usepackage{array}
\usepackage{xcolor}
\usepackage[italian]{babel}
\usepackage[utf8]{inputenc}
\usepackage{verbatim}
\usepackage[active]{srcltx}
\usepackage{url}
\usepackage[T1]{fontenc}
\usepackage{mathtools} 
\usepackage{subfig}
\usepackage{cite}
\usepackage{microtype} 
\usepackage{bm}
\usepackage{dsfont}

\title{\LARGE \bf
Parameter Identification of a Differentiable Human Arm Musculoskeletal Model without Deep Muscle EMG Reconstruction
}

\author{Philip Sanderink, Yingfan Zhou, Shuzhen Luo, Cheng Fang
}

\begin{document}

\maketitle
\thispagestyle{empty}
\pagestyle{empty}

\begin{abstract}

Accurate parameter identification of a subject-specific human musculoskeletal model is crucial to the development of safe and reliable physically collaborative robotic systems, for instance, assistive exoskeletons. Electromyography (EMG)-based parameter identification methods have demonstrated promising performance for personalized musculoskeletal modeling, whereas their applicability is limited by the difficulty of measuring deep muscle EMGs invasively. Although several strategies have been proposed to reconstruct deep muscle EMGs or activations for parameter identification, their reliability and robustness are limited by assumptions about the deep muscle behavior. In this work, we proposed an approach to simultaneously identify the bone and superficial muscle parameters of a human arm musculoskeletal model without reconstructing the deep muscle EMGs. This is achieved by only using the least-squares solution of the deep muscle forces to calculate a loss gradient with respect to the model parameters for identifying them in a framework of differentiable optimization. The results of extensive comparative simulations manifested that our proposed method can achieve comparable estimation accuracy compared to a similar method, but with all the muscle EMGs available.

\end{abstract}

\section{Introduction}

In recent years, a variety of robots have gradually been integrated into human work and daily life. For better safety and performance, human musculoskeletal modeling, as an essential component, has also been actively incorporated into diverse physical human-robot interaction scenarios like ergonomic co-manipulation with collaborative robots, rehabilitation therapy planning with assistive exoskeleton devices, and the restoration of mobility with powered prosthetic limbs \cite{fang2023human, seth2018opensim, wang2022myosim, peternel2020human}. Reactions to the same robot behavior may vary across different individuals. Personalizing a human musculoskeletal model by identifying individualized model parameters is the first crucial step in simulating subject-specific dynamic behavior in a simulator, which facilitates the design of a robot controller prior to real-world implementation. In addition, accurate estimation of subject-specific parameters can also enable convenient evaluation of muscle and tendon quality, such as, muscle strength and tendon stiffness, with wide applications across sports science, rehabilitation engineering, and medical diagnostics \cite{barbat2012assess, wu2016subject, uchida2021biomechanics}.

\begin{figure}[!t]
    \centering
    \includegraphics[trim = 0mm 0mm 0mm 0mm, clip, width=1\linewidth]{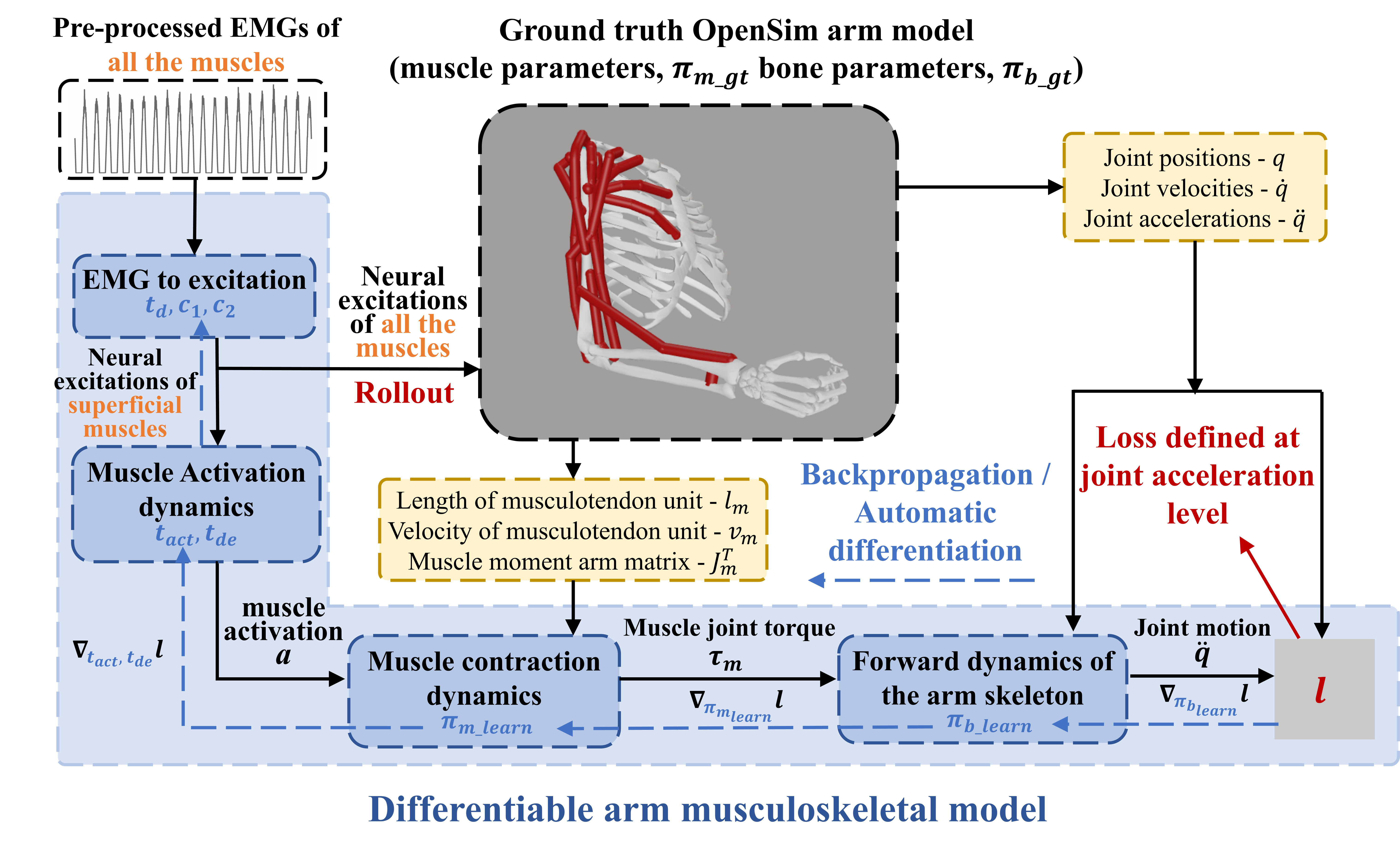}
    \caption{A differentiable human arm musculoskeletal model for simultaneous identification of the bone and muscle parameters without deep muscle EMG reconstruction.}
    \label{fig:model}
    \vspace{-0.3cm}
\end{figure}

Typically, a subject-specific musculoskeletal model is built by scaling a template model, with base parameters obtained or estimated from cadaver-based statistical datasets representing a given population \cite{andersen2021introduction}. The scaling factors are calculated based on the recorded motion data of an individual and they are used to scale both bone dynamic parameters and muscle parameters. Although it has been widely demonstrated that the inaccuracy of bone parameter estimates has a lesser impact on the inaccuracy of dynamic behavior than the inaccuracy of kinematic parameters \cite{muller2017uncertainty}, the muscle parameters do have a significant effect \cite{winby2008evaluation, modenese2016estimation}.  
To improve muscle parameter estimation, individual EMG signals can be utilized \cite{sartori2012emg, falisse2016emg}. Well-preprocessed EMGs can be used to drive a musculoskeletal model to calculate the joint torque and the resulting motion, which can be used to compute a loss by comparing it against the observed motion. The muscle and bone parameters can thus be estimated by solving a non-linear optimization problem that minimizes the loss \cite{venture2005identifying,hayashibe2011muscle, durandau2017robust, zhou2025diff}. However, activities of superficial muscles can be easily measured by surface EMG sensors, whereas those of deep muscles cannot \cite{peter2019comparing}. In practice, deep muscles can be measured using fine-wire electrodes, but such measurements are invasive and require specialized expertise; therefore, they are impractical for daily use or large-scale applications \cite{al2023electromyography}. Such a limitation raises a question: \textit{Can some of the musculoskeletal model parameters still be identified without access to the EMG recordings of deep muscles?}

This research makes two primary contributions:

1) We propose simultaneously identifying the superficial muscle and bone parameters of a human arm musculoskeletal model without reconstruction of deep muscle EMG signals using differentiable optimization for the first time.

2) We present extensive simulations with a high-fidelity human arm musculoskeletal model (11 superficial muscles and 5 deep muscles) that demonstrate our proposed approach can achieve good estimation results.

\section{Related works}
\vspace{-0.1cm}
EMG-based musculoskeletal modeling approach has been widely employed for musculoskeletal model calibration to identify muscle parameters (e.g., optimal fiber length and maximum isometric force) and/or rigid bone dynamic parameters (e.g., mass and center of mass). The parameter estimation is typically formulated as an optimization problem with a loss function defined either at the joint torque or joint acceleration level. If the experimental EMG data of all the muscles of interest can be obtained, and skeletal dynamic parameters are available, either from statistical mass distribution models \cite{winter2009biomechanics} or from scaling techniques driven by kinematic data \cite{fang2018real}, internal joint torques can be derived through inverse dynamics of the rigid skeleton with measured motion as the inputs. Then, muscle parameters can be estimated by solving the optimization problem with the loss defined between the derived joint torques and the predicted ones calculated from muscle activation and contraction dynamics models with the neural excitations as the input obtained from processed EMG recordings  \cite{venture2005identifying, hayashibe2011muscle, durandau2017robust, zhou2025diff}. Due to the strong non-linearity and non-convexity inherent in muscle dynamics, this optimization problem is typically addressed through global optimization algorithms, such as simulated annealing \cite{goffe1994global} or particle swarm algorithms \cite{schutte2005evaluation}. However, any inaccuracies of the bone parameter estimates would propagate to the calculated joint torques and ultimately compromise the accuracy of the identified muscle parameters \cite{rao2006influence}. To tackle this issue, a differential musculoskeletal model was proposed to simultaneously estimate the muscle and bone parameters, which demonstrated better results against the global optimization methods in a simulation study \cite{zhou2025diff}.


However, in real experiments, it is not practical to easily acquire EMG signals from deep muscles. To cope with this limitation, the EMG-assisted strategy allows for synthesizing unmeasurable muscle EMG signals and, meanwhile, adjusting recorded observable EMG signals to reduce the discrepancy between the predicted joint torques and the measured ones \cite{pizzolato2015ceinms}. However, introducing synthesized or altered EMG signals into the calibration may compromise its fidelity and yield physiologically unreliable estimates. Alternatively, muscle synergy extrapolation (MSE) has been also widely used to estimate EMG data from deep or missing muscles. Muscle synergy refers to the coordinated activation of groups of muscles that act together as functional units to produce movement \cite{steele2013number}. With MSE, recorded surface EMG signals are projected onto a reduced subspace represented by time-varying synergy excitations as the basis functions and time-invariant weight vectors. The extracted basis synergy patterns are subsequently utilized to reconstruct the EMG activity of deep muscles, which, together with the measured EMG signals from surface muscles, are used to drive EMG-based musculoskeletal models for parameter identification \cite{ao2024comparison, ao2022emg, sartori2013musculoskeletal, rabbi2024muscle}. In particular, principal component analysis and non-negative matrix factorization are the most widely adopted approaches for synergy extraction \cite{lambert2017identifying}. Although MSE strategy yields physiologically consistent muscle excitation reconstruction, the reliability of the reconstructed signals depends heavily on the specific extraction method employed and the number of synergies to be selected, which can substantially compromise the accuracy of deep muscle reconstruction and consequently hinder the reliable identification of model parameters \cite{turpin2021improve}. In recent years, advances in deep neural networks have introduced new perspectives for EMG-based parameter identification. Particularly, physics-informed neural networks enhance musculoskeletal model parameter identification by embedding a physics-based model into a data-driven learning paradigm, thereby improving interpretability and accuracy for parameter identification even with limited or unlabeled data \cite{taneja2022feature, zhang2022boosting, ma2024physics}. Recently, to address the problem of the lack of deep muscle EMGs, a neural network-based musculoskeletal model was proposed to reconstruct deep muscle EMGs from motion data while integrating surface EMGs to ensure physiological consistency and simultaneously enable musculoskeletal parameter identification \cite{kumar2025deep}. However, the identified parameters were not tested in new datasets with movement and conditions different from the training data. 


\section{Our approach}
Since the invasive and accurate reconstruction of the deep muscle EMGs is difficult and different combinations of EMGs might generate the same observed movement (redundancy in the muscle recruitment), we propose to identify the model parameters of a human arm musculoskeletal model without the EMG reconstruction of the deep unmeasurable muscles.

\subsection{Differentiable optimization}

There is an emerging paradigm of solving the parameter estimation in sim-2-real gap \cite{lutter2021differentiable, granados2022model, wang2023real2sim2real} in robotics by using differential optimization/simulation \cite{degrave2019differentiable, newbury2024review}, where the gradients of the loss relative to the model parameters are computed by automatic differentiation \cite{paszke2017automatic} on a computational graph reflecting the topological dependency between a large number of intermediate variables and parameters of a model, and the gradients are used to minimize the loss to find the optimal model parameters. The power of the differentiable optimization for simultaneously identifying a large number of bone and muscle parameters of a musculoskeletal model has been demonstrated in \cite{zhou2025diff} recently. 

\subsection{Differentiable arm musculoskeletal model}

To have better model realism as preparation for the future real experiments, an OpenSim human arm musculoskeletal model, ``\textit{MoBL-ARMS Dynamic Upper Limb}''\cite{saul2015benchmarking}, instead of a simplified MyoSuite model (used in \cite{zhou2025diff}), is adopted as a reference model to generate ground truth training data. A differentiable version of this OpenSim model is developed outside the OpenSim to carry out the differentiable optimization (shown in Fig. \ref{fig:model}) where the bone and muscle parameters are allowed to be changed and optimized to minimize a loss (error) of a variable, e.g., joint acceleration. In our differentiable arm model, the shoulder and elbow joints are movable while the wrist and finger joints are locked. The shoulder consists of three independent joint Degrees of Freedom (DoFs): flexion-extension (\textit{Sh\_Fle\_Ext}), abduction-adduction (\textit{Sh\_Abd\_Add}), and medial-lateral rotation (\textit{Sh\_Med\_Lat}) DoFs, while the elbow is comprised of two independent joint DoFs: flexion-extension (\textit{El\_Fle\_Ext}) and pronation-supination (\textit{El\_Pro\_Sup}) DoFs. The differentiable model includes eleven superficial muscles and five deep muscles, which are important for the movement generation of the shoulder and elbow joints. The superficial muscles include: the superior (\textit{PECM}1), middle (\textit{PECM}2), and inferior (\textit{PECM}3) portions of the Pectoralis Major (Prime Mover, PM, of \textit{Sh\_Fle\_Ext} and \textit{Sh\_Abd\_Add}); the anterior (\textit{DELT}1), lateral (\textit{DELT}2), posterior (\textit{DELT}3) portions of the Deltoid (PM of \textit{Sh\_Fle\_Ext} and \textit{Sh\_Abd\_Add}); the long (\textit{BIClong}) and short portions (\textit{BICshort}) of the Biceps (PM of \textit{El\_Fle\_Ext}); the long (\textit{TRIlong}) and lateral (\textit{TRIlat}) portions of the Triceps (PM of \textit{El\_Fle\_Ext}), and Brachioradialis (\textit{BRD}). The deep muscles are: the Infraspinatus (\textit{INFSP}, PM of \textit{Sh\_Med\_Lat}), Subscapularis (\textit{SUBSC}, PM of \textit{Sh\_Med\_Lat}), the medial portion of the Triceps (\textit{TRImed}, PM of \textit{El\_Fle\_Ext}), Brachialis (\textit{BRA}, PM of \textit{El\_Fle\_Ext}), and the Pronator quadratus (\textit{PQ}, PM of \textit{El\_Pro\_Sup}) muscles \cite{marieb2007human}. 

In contrast to the differentiable arm model used in \cite{zhou2025diff}, to improve the realism of the model and make it ready for practical use in the real experiments, deep PM muscles are modeled but their EMGs are not assumed to be available, the EMG to muscle neural excitation dynamics and muscle activation dynamics are added, and constrained arm model is also considered in this work.    
\vspace{-0.1cm}

\subsection{Constrained arm model}

In the OpenSim model, there are eleven dependent DoFs whose values depend on the three independent DoFs of the shoulder joint. These dependencies are formulated as kinematic constraints in the dynamic system of the arm skeleton to incorporate, for instance, the effect of the topological loops created by sets of bones and joints. The constraints also introduce constraint torques to the equation of motion of the arm \cite{sherman2011simbody}:
\begin{equation}
    \bm{M}\ddot{\bm{q}} + \bm{\tau}_{bias} + \bm{G}^T\bm{\lambda} = \bm{J}_{m}^T\bm{f}_{m},
    \label{eq:eom}
\end{equation}
\begin{equation}
    \bm{G}\ddot{\bm{q}} = \bm{b},
    \label{eq:eom_con}
\end{equation}
where $\bm{M}_{_{16\times16}}$ is the inertia matrix, $\ddot{\bm{q}}_{_{16\times1}}$denotes the accelerations of all the 16 independent and dependent joint DoFs, $\bm{\tau}_{bias_{16\times1}}$ indicates the bias torque because of the Coriolis and centrifugal effects and the gravity. $\bm{G}_{_{11\times16}}$ and $\bm{b}_{_{11\times1}}$ in (\ref{eq:eom_con}) are constant matrix and vector which are used to describe the constraints, and a vector of unknown Lagrange multipliers, $\bm{\lambda}_{_{11\times1}}$, is mapped to the constraint torques through $\bm{G}^T$ in (\ref{eq:eom}). $\bm{f}_{m_{16\times1}}$ means the forces produced by all the 16 muscles, and $\bm{J}_{m_{ 16\times16}}$ is a muscle Jacobian, in which each element is a moment arm of a muscle with respect to a joint DoF calculated by the partial derivative of the length of a musculotendon unit relative to a DoF \cite{fang2017online}. $\bm{J}_{m}^T\bm{f}_{m}$ represents the torques $\bm{\tau}_{m}$ applied by all the muscles.

\subsection{Addressing deep muscle behavior}

\begin{figure}[!t]
     \centering
     \includegraphics[trim = 0mm 0mm 0mm 0mm, clip, width=1\linewidth]{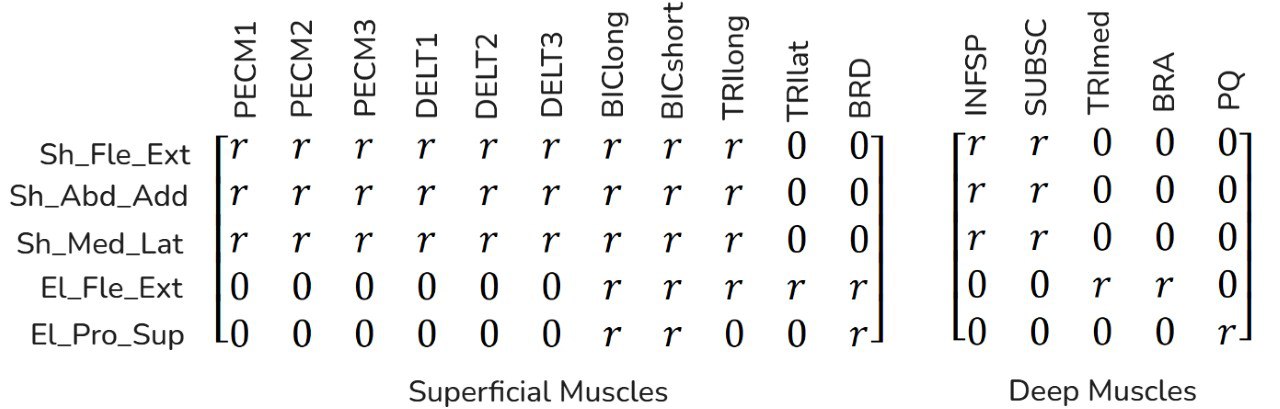}
     \caption{Moment arms of different superficial and deep muscles of a human arm musculoskeletal model. $r$ means a muscle spans over a joint and therefore has a moment arm with respect to the joint.}
     \label{fig:moment_arm}
\end{figure}

Let's examine how the deep muscles would affect the dynamics of the arm skeleton, focusing on the five independent DoFs. The deep muscles, \textit{INFSP} and \textit{SUBSC}, span over the three DoFs of the shoulder, \textit{TRImed} and \textit{BRA} only affect the \textit{EL\_Fle\_Ext} of the elbow, and \textit{PQ} only causes joint torque on the \textit{EL\_Pro\_Sup} of the elbow throughout most of the range of motion. The contribution of the deep muscles to the joint torques can be described by the transpose of the deep muscle Jacobian, $\bm{J}_{md}^T$\footnote{The real dimension of $\bm{J}_{md}^T$ is $16\times5$ where the moment arms of the deep muscles over the 11 dependent DoFs are zeros. The dependent DoFs are ignored in Fig. \ref{fig:moment_arm} for better expression clarity.}, shown on the right side of Fig. \ref{fig:moment_arm}. It can be easily seen that the $3\times2$ non-zero muscle moment arm submatrix in the upper left corner of $\bm{J}_{md}^T$ renders an overdetermined system of linear equations in terms of the joint torques of the three shoulder DoFs which have to be provided by the \textit{INFSP} and \textit{SUBSC}, while the $2\times3$ moment arm submatrix in the lower right corner of $\bm{J}_{md}^T$ makes an underdetermined system of linear equations in terms of the required joint torques of the two elbow DoFs for the \textit{TRImed}, \textit{BRA}, and \textit{PQ}. These two decoupled submatrices are the important matrix properties we will utilize to estimate the model parameters.

We propose to treat the deep muscle behavior as a black box, and only care about the output of the box, which is the muscle force $\bm{f}_{m}$, without modeling the internal muscle dynamic behavior, i.e., without reconstructing the deep muscle EMGs or optimizing their muscle parameters. By virtue of the overdetermined system, all the bone parameters and the muscle model parameters of the muscles that would affect the three shoulder DoFs (i.e., the \textit{PECM}1, \textit{PECM}2, \textit{PECM}3, \textit{DELT}1, \textit{DELT}2, \textit{DELT}3, \textit{BIClong}, \textit{BICshort}, and \textit{TRIlong} shown in Fig. \ref{fig:moment_arm}) can be estimated if we define the loss as the error between the missing deep muscle torques and the provided torques calculated by the least-squares solution of the muscle forces of the the \textit{INFSP} and \textit{SUBSC}. This is because if the model parameter estimates are inaccurate, they would generate wrong missing deep muscle torques of the three shoulder DoFs, which cannot be perfectly compensated by the best possible solution of the deep muscle forces unless the estimates are at the ground truth. The residual torque error makes the gradient of the loss with respect to the model parameters non-zero, which can guide the optimization to find the right model parameters. On the other hand, due to the underdetermined system, the muscle parameters of the superficial muscles, which only affect the two elbow DoFs (i.e., \textit{TRIlat} and \textit{BRD} shown in Fig. \ref{fig:moment_arm}) cannot be estimated because any missing wrong deep muscle torques of the two elbow DoFs caused by inaccurate muscle parameters can be always well compensated by infinite number of solutions of the forces provided by the \textit{TRImed}, \textit{BRA}, and \textit{PQ}. Therefore, the loss gradient would vanish. 

\subsection{Loss calculation}

The loss caused by the proposed least-squares solution of the deep muscle forces can be calculated based on the constrained arm model ((\ref{eq:eom}) and (\ref{eq:eom_con})) with known motion ($\bm{q}, \dot{\bm{q}}, \ddot{\bm{q}}$) and estimated $\bm{M}$ and $\bm{\tau}_{bias}$. Let's assume,  
\begin{equation}
    \ddot{\bm{q}} = \ddot{\bm{q}}_u - \ddot{\bm{q}}_c,
    \label{eq:q}
\end{equation}
\begin{equation}
    \ddot{\bm{q}}_u = \bm{M}^{-1} (\bm{\tau}_{m} -\bm{\tau}_{bias}),
    \label{eq:q_u}
\end{equation}
where $\ddot{\bm{q}}_u$ and $\ddot{\bm{q}}_c$ are the joint accelerations of the corresponding unconstrained system and the offset caused by the constraints, respectively. Substituting (\ref{eq:q}) into (\ref{eq:eom_con}), we have,
\begin{equation}
    \bm{G}\ddot{\bm{q}}_c = \bm{G}\ddot{\bm{q}}_u -\bm{b}.
    \label{eq:gq_c}
\end{equation}
In the meantime, after substituting (\ref{eq:q}) and (\ref{eq:q_u}) into (\ref{eq:eom}), it can be seen that,
\begin{equation}
    \bm{M}\ddot{\bm{q}}_c = \bm{G}^T\bm{\lambda}.
    \label{eq:qc}
\end{equation}
To solve for $\ddot{\bm{q}}_c$ with (\ref{eq:gq_c}) alone, there exists an infinite number of solutions due to the underdetermined system in (\ref{eq:gq_c}). However, the solution of $\ddot{\bm{q}}_c$ must be within the image space of $ \bm{M}^{-1}\bm{G}^T $ shown in (\ref{eq:qc}), therefore, a pseudoinverse with a weighting matrix of $\bm{M}$ must be used to solve for $\ddot{\bm{q}}_c$: 
\begin{equation}
    \ddot{\bm{q}}_c = \bm{G}^{M+}(\bm{G}\ddot{\bm{q}}_u -\bm{b}) = \bm{M}^{-1}\bm{G}^T(\bm{G}\bm{M}^{-1}\bm{G}^T)^{-1}(\bm{G}\ddot{\bm{q}}_u -\bm{b}),
    \label{eq:qc_solution}
\end{equation}
where the invertibility of $\bm{G}\bm{M}^{-1}\bm{G}^T$ is checked to ensure the existence of the weighted pseudoinverse for all the data points during the optimization. Accordingly, the Lagrange multiplier vector $\bm{\lambda}$ can be obtained with (\ref{eq:qc}) and (\ref{eq:qc_solution}):
\begin{equation}
    \bm{\lambda} = (\bm{G}\bm{M}^{-1}\bm{G}^T)^{-1}(\bm{G}\ddot{\bm{q}}_u -\bm{b}).
    \label{eq:lambda}
\end{equation}
By replacing $\ddot{\bm{q}}_u$ in (\ref{eq:lambda}) with (\ref{eq:q_u}), and substituting the expression of $\bm{\lambda}$ back to (\ref{eq:eom}), we can get:
\begin{equation}
    \begin{aligned}
        &\bm{M}^{-1}(\bm{\tau}_{m}-\bm{\tau}_{bias})=\\&\bm{M}^{-1}\bm{G}^T(\bm{G}\bm{M}^{-1}\bm{G}^T)^{-1}(\bm{G}\bm{M}^{-1}(\bm{\tau}_{m}-\bm{\tau}_{bias})-b) +\ddot{\bm{q}}
    \end{aligned}\label{eq:tau}
\end{equation}
To simplify (\ref{eq:tau}), we define the matrices and vector: 
\begin{equation}
\begin{aligned}
    \bm{A}(\bm{q}, \bm{\pi}_b) &= (\bm{I} - \bm{G}^{M+}\bm{G})\bm{M}^{-1},
    \\
    \bm{c}(\bm{q}, \dot{\bm{q}}, \ddot{\bm{q}}, \bm{\pi}_b) &= \ddot{\bm{q}} + \bm{A}\bm{\tau}_{bias} - \bm{G}^{M+}\bm{b},
    \label{eq:ABC}
\end{aligned}
\end{equation}
where $\bm{I}$ is an identity matrix and the motion variables and dynamic parameters of bones $\bm{\pi}_b$ in the parentheses show the dependency of $\bm{A}$ and $\bm{c}$. (\ref{eq:tau}) can be then rearranged into: 
\begin{equation}
    \bm{A}(\bm{q}, \bm{\pi}_b)\bm{\tau}_{m} = \bm{c}(\bm{q}, \dot{\bm{q}}, \ddot{\bm{q}}, \bm{\pi}_b),
    \label{eq:tau_m}
\end{equation}
where it is worth mentioning that $\bm{A}$ transforms the muscle joint torque $\bm{\tau}_{m}$ to a joint acceleration to leverage on the availability of the ground truth $\ddot{\bm{q}}$ in $\bm{c}$ for the loss calculation.

With the respective contribution of the superficial and deep muscles to $\bm{\tau}_{m}$, (\ref{eq:tau_m}) can be expressed as:
\begin{equation}
    \bm{A}(\bm{J}_{ms}^T(\bm{q})\bm{f}_{ms}(\bm{q}, \dot{\bm{q}}, \bm{e}_{ms}, \bm{\pi}_{ms}) + \bm{J}_{md}^T(\bm{q})\bm{f}_{md}) = \bm{c},
    \label{eq:tau_m2}
\end{equation}
where $\bm{J}_{ms}$ and $\bm{f}_{ms}$ are the muscle Jacobian and force vector of the superficial muscles. $\bm{J}_{ms}$ and $\bm{J}_{md}$ depend on the arm configuration $\bm{q}$, which can be obtained from the measured arm movement, and $\bm{f}_{ms}$ can be computed based on a muscle model (introduced in \ref{sec:muc_con}) with measured and preprocessed EMGs $\bm{e}_{ms}$ and the muscle model parameters $\bm{\pi}_{ms}$. Therefore, the least-squares solution of the unknown deep muscle forces $\bm{f}_{md}$ can be calculated using the pseudoinverse $(\cdot)^{+}$ as:
\begin{equation}
    \bm{f}_{md} = (\bm{A}\bm{J}_{md}^T)^{+} (\bm{c} - \bm{A}\bm{J}_{ms}^T\bm{f}_{ms}),
    \label{eq:tau_m3}
\end{equation}

Based on (\ref{eq:tau_m2}) and (\ref{eq:tau_m3}), the parameter identification problem is formulated as an optimization problem with the loss function defined as follows:
\begin{equation}
\begin{aligned}
    l &= \frac{\sum_{t=1}^{T} \left\| (\bm{I} - \bm{A}\bm{J}_{md}^T(\bm{A}\bm{J}_{md}^T)^{+}) \bm{d}_t (\bm{\pi}_b, \bm{\pi}_{ms}) \right\|_2^2}{T\times var(\bm{d}_t)}, \\
    \bm{d} &= \bm{c} - \bm{A}\bm{J}_{ms}^T\bm{f}_{ms},
    \label{eq:loss}
\end{aligned}
\end{equation}
which is the normalized mean squared error due to the least-squares solution in (\ref{eq:tau_m3}). \( T \) represents the number of trajectory points in the dataset, and \( var(\bm{d}_t)\) denotes the variance of $\bm{d}_t$ with the initial guess of all the parameters. The model parameters to be identified in the optimization problem include ten dynamic parameters (mass, center of mass, and six parameters of the inertia matrix) for each of the \textit{humerus}, \textit{ulna}, and \textit{radius} ($\bm{\pi}_{b}\in\mathbb{R}^{30}$), and five parameters for each of the nine identifiable superficial muscles, and five other parameters common to all the muscles ($\bm{\pi}_{ms}\in\mathbb{R}^{50}$), which will be introduced shortly in the following subsection.

\subsection{Improving model realism}

\subsubsection{Muscle contraction dynamics} \label{sec:muc_con}
A musculotendon unit is modeled by a Hill-type model \cite{zajac1989muscle}, in which a muscle is composed of an active contractile element in parallel to a passive elastic element, which is connected in series at a pennation angle ($\phi$) to a tendon modeled as a passive elastic nonlinear spring. The force production of this system is governed by the muscle contraction dynamics, which converts the muscle activation $a$ to the musculotendon force $f_{mt}$\footnote{$f_{mt}$ represents each of the elements in the muscle force vector $\bm{f}_m$ in (\ref{eq:eom}), the subscript $t$ here indicates the fact that the muscle force is delivered to the bone through a tendon for causing a joint torque.} as:
\begin{align}
f_{mt} &= cos(\phi) f^o_m(f^L(\tilde{l}_m)f^V(\tilde{v}_m)a+f^{PE}(\tilde{l}_m)), \label{eq:muscle} \\
f_{mt} &= f^o_m(f^{T}(\tilde{l}_t)), \label{eq:tendon}
\end{align}
where both (\ref{eq:muscle}) and (\ref{eq:tendon}) can be used to calculate the force $f_{mt}$ because of the force equilibrium between the muscle and the tendon \cite{uchida2021biomechanics}. $f^L \text{,}~ f^V \text{,}~ f^{PE}\text{,}~ f^T$ are four generic, normalized, and time-invariant force curves for all the musculotendon units. Their inputs: the muscle length $l_m$, muscle velocity $v_m$, and tendon length $l_t$, are normalized by the optimal muscle fiber length ($\tilde{l}_m = l_m/l_m^o$), maximum muscle contraction velocity ($\tilde{v}_m = v_m/v_m^{max}$), and tendon slack length ($\tilde{l}_t = l_t/l_t^s$), respectively. Their outputs are normalized by the maximum muscle isometric force $f_m^o$. $f_m^o \text{,}~ l_m^o \text{,}~ v_m^{max}\text{,}~ \phi_o(\phi \text{~when~} l_m = l_m^o)\text{, and}~ l_t^s$ are the five musculotendon-specific parameters to be identified for each of the nine identifiable superficial muscles. 

In the MyoSuite model used in \cite{zhou2025diff}, the tendon is assumed to be rigid. Therefore, $l_t^s$ does not exist, and the pennation angle is not modeled either. In addition, the four force curves are simplified in the MyoSuite model. However, $l_t^s$, $\phi_o$, and the curves are very important for the force-generating behavior of a musculotendon unit \cite{winby2008evaluation, modenese2016estimation}. In OpenSim, quintic Bézier splines are used to model these curves, allowing for accurate fitting of the experimental data \cite{millard2013flexing}. These splines are parametric equations where the parameter needs to be first calculated based on the input to be able to compute the corresponding output. An iterative numerical algorithm, such as the Newton-Raphson method used in OpenSim, is required to find the solution of the fifth-order polynomial equation for the parameter, which is slow when integrated into a large-scale differentiable optimization framework. To make the optimization more efficient, a differentiable piecewise function consisting of between 8 and 15 polynomials connected by optimized line segments is developed to fit each of the four OpenSim force curves. Since the polynomials directly express the relation between the input and output, they allow for efficient calculation of the derivative of the output with respect to the input as part of the chain of automatic differentiation. In the full differentiable optimization pipeline, (\ref{eq:muscle}) is first used to optimize the four muscle-related parameters, i.e., $f_m^o \text{,}~ l_m^o \text{,}~ v_m^{max}\text{, and}~ \phi_o$, and when their good estimates are found (loss is low), (\ref{eq:tendon}) is then used to optimize $l_t^s$.

\subsubsection{Muscle activation dynamics}
As shown in Fig. \ref{fig:model}, muscle activation dynamics is an upstream module with respect to the muscle contraction dynamics. It models how the neural excitations $u$ generated by the alpha motor neurons can affect the activation level $a$ of their innervated muscles as \cite{uchida2021biomechanics}:
\begin{equation}
    \dot{a}=\frac{u-a}{\mathfrak{t}(u,a)}
    \label{eq:act_dyn}
\end{equation}
\begin{equation}
    \mathfrak{t}(u,a)=
    \begin{cases}
        \mathfrak{t}_{act}(0.5+1.5a)   & u > a\\
        \mathfrak{t}_{de}/(0.5 + 1.5a) & u \leq a\\
    \end{cases} 
\end{equation}
where $\mathfrak{t}_{act}$ and $\mathfrak{t}_{de}$ are the activation and deactivation time constants, which affect how fast the muscles are activated and deactivated by $u$. These two parameters are assumed to be the same for different muscles, which need to be identified in the optimization.

At each iteration during the optimization, when the parameters are updated, the time integration of (\ref{eq:act_dyn}) for getting the full profile of $a$ is time-consuming when the number of time steps is large. Since $a,u\in[0,1]$, it is guaranteed that $\mathfrak{t}>0$, which ensures the solution of the exponentially decaying system (\ref{eq:act_dyn}) can converge to the excitation $u$ over time. By leveraging the guaranteed convergence, the full trajectory is divided into smaller segments, which are all assumed to have an initial $a$ of 0. An explicit Runga-Kutta method is used to solve for every segment of the trajectory in parallel to speed up the time integration. This would result in inaccurate activations at the beginning of each segment (except the first one) due to the wrong initial guess, but the activations then quickly converge to the ground truth. The errors at the beginning of each segment are eliminated by redoing the time integration for the first few time steps of each segment, using the accurate final activation of the previous segment as the initial $a$ in a second pass.

\subsubsection{EMG-to-excitation dynamics}
To capture the electromechanical time delay $t_d$ from the onset of the EMG to the onset of the resulting muscle force production, an EMG-to-excitation dynamics \cite{lloyd2003emg} is used as an upstream module for the muscle activation dynamics. It converts the preprocessed EMG signals $e$ to the neural excitation $u$, which is modeled by a recursive filter \cite{lloyd2003emg}:
\begin{equation}
    u(t) = \alpha e(t-d) - \beta_1 u(t-1) - \beta_2 u(t-2),
    \label{eq:u_filter_tf}
\end{equation}
where $d$ is the time delay, $\alpha,\beta_1,\beta_2$ are filter coefficients. To ensure the filter has unit gain and stability, the following relations must hold:
\begin{equation}
\begin{aligned}
    \alpha &= 1+\beta_1+\beta_2, \\
    \beta_1 &= c_1 + c_2, \\
    \beta_2 &= c_1 \cdot c_2, \\
    |c_1|,|c_2|&<1. 
\end{aligned}
\end{equation}

$c_1$, $c_2$, and $d$ are subject-specific \cite{lloyd2003emg} and are therefore considered learnable parameters common to all the muscles in our model. Directly making $d$ learnable is not feasible, since it is a non-continuous index (an integer representing multiples of the sampling period) with respect to which the derivative of $u$ doesn't exist. To solve this, we define the continuous time delay $t_d$ as the learnable parameter, and design a differentiable function to map $t_d$ to the continuous coefficients, $\gamma$ and $1-\gamma$, of the two neighboring time indices, $d_l$ and $d_h$, as: 
\begin{equation}
\begin{aligned}
    u(t) =~& \alpha [ \gamma e(t-d_l) + (1-\gamma) e(t-d_h)] \\
        & - \beta_1 u(t-1) - \beta_2 u(t-2), \\
        d_h =~& \text{ceil}(t_d) \\
        d_l =~& \text{floor}(t_d) \\
        \gamma =~&(1-t_d+d_l)
\end{aligned}
\end{equation}

\section{Comparative simulations}

To the best knowledge of the authors, there exists no method for parameter identification without the reconstruction of deep muscle EMGs (in the cases where the effect of the deep PM muscles cannot be ignored). To examine how well the model parameters can be estimated without deep muscle EMG reconstruction, our proposed method (M1) was compared against the simultaneous optimization method (M2), where all the muscle EMGs were assumed to be available (as the best possible case of deep muscle EMG reconstruction).

The loss calculation, initial conditions, and differentiable optimization pipeline of M1 and M2 are identical, except for the availability of deep muscle EMGs. The PyTorch automatic differentiation tool \cite{paszke2017automatic} was used to calculate the loss gradient and execute the differentiable optimization for both M1 and M2, and only the identifiable model parameters of M1 were used for the comparison between the two methods.

\subsection{Comparison settings}

\begin{figure}
    \centering
    \includegraphics[width=0.8\linewidth]{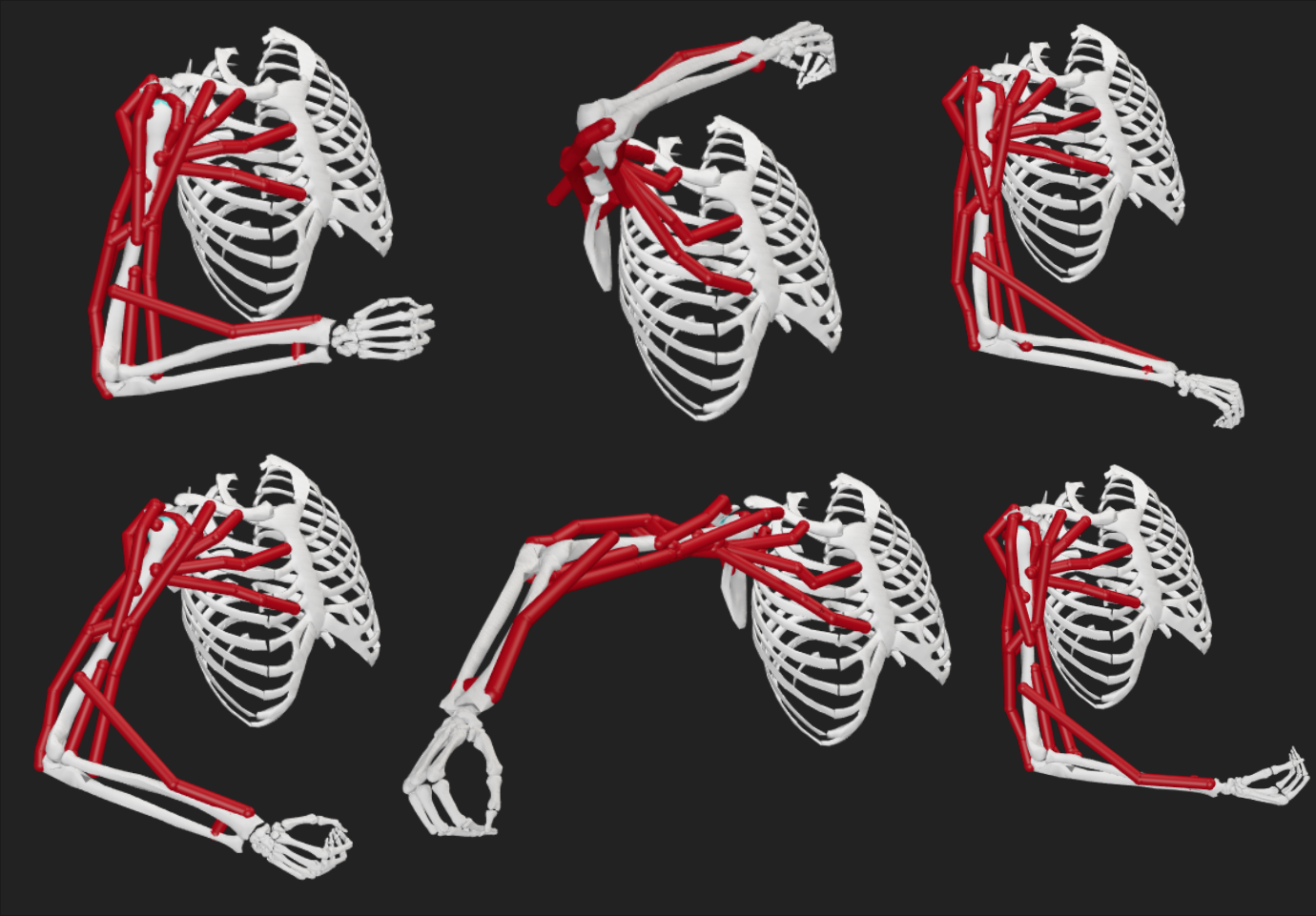}
    \caption{Snapshots of the OpenSim arm configurations taken at three different times for each of two arm movements used in the training (each row represents one arm movement).}
    \label{fig:settings:opensimarm}
\end{figure}

Training datasets of two different arm movements were generated by driving the ground truth arm model in OpenSim through the forward dynamics with the neural excitations of all the muscles as the input calculated from the synthesized irregular EMG signals, shown in Fig. \ref{fig:model}. The excitations of only the superficial muscles were fed into our differentiable arm model for the differentiable optimization in M1, while all the muscle excitations were provided in M2. At each time step, the collected data comprised joint positions, velocities, and accelerations, as well as the muscle lengths, velocities, and moment arms. Fig. \ref{fig:settings:opensimarm} shows the snapshots taken at three different times for each of the two arm movements. 
\vspace{-0.1cm}



\subsection{Evaluation criteria}


Three criteria, \( Cr_{1}\), \( Cr_{2}\), and $Cr_2^m$, were used for the evaluation of the parameter estimation accuracy. \( Cr_{1}\) describes how far a solution is from the ground truth:
\begin{equation}
    Cr_{1}=\frac{\|\hat{\bm{\pi}}-\bm{\pi}_{gt}\|_2}{\|\bm{\pi}_{gt}\|_2} \times 100\%,
\end{equation} 
and \( Cr_{2}\) is the mean of the percentage errors of
all the estimated parameters, eliminating the effect of different units: 
\begin{equation}
    Cr_{2}=\frac{1}{n}\sum_{i=1}^{n}\frac{|\hat{\pi}_i-\pi_{gt_{i}}|}{|\pi_{gt_{i}}|}\times 100\%,
\end{equation}
and $Cr_2^m$ refers to the mean of the percentage errors of only estimated muscle parameters.

\vspace{-0.1cm}

\subsection{Results}

\begin{figure}[!t]
    \centering
    \includegraphics[trim = 10mm 0mm 5mm 10mm, clip, width=\linewidth]{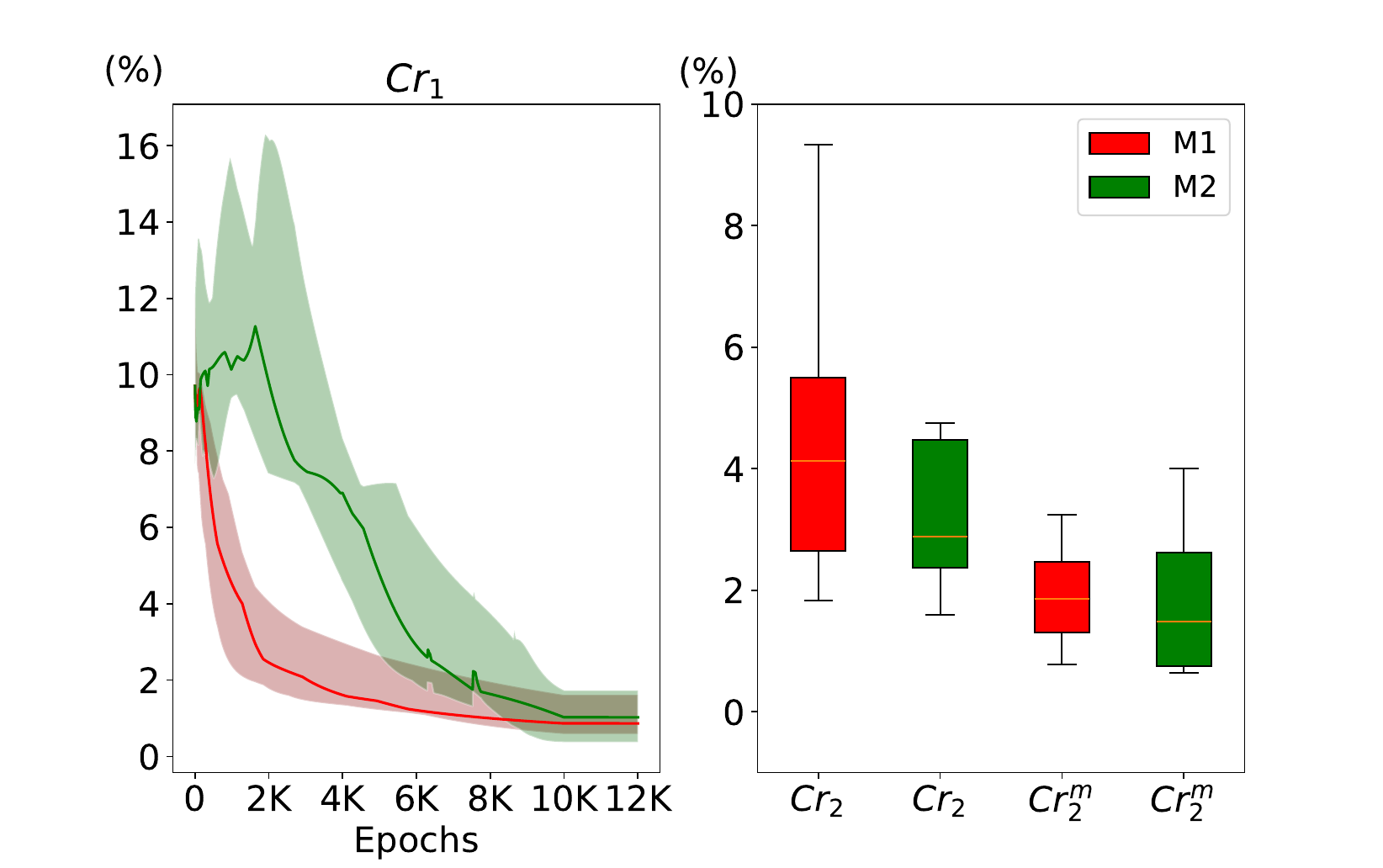}
    \caption{Distance error curves in percentage ($Cr_1$) of 10 runs over 12K iterations of M1: differential optimization without deep muscle EMGs and M2: differential optimization with deep muscle EMGs. The solid line denotes the median, while the upper and lower boundaries of the shaded area represent upper and lower quartiles, respectively (left). Boxplots of final parameter estimate mean error $Cr_2$ and final muscle parameter mean error $Cr_2^m$ (right).}
    \label{fig:res:C1C2} 
\end{figure}
\vspace{-0.1cm}    

\begin{figure*}[!t] \centering
	{\includegraphics[width=2\columnwidth]{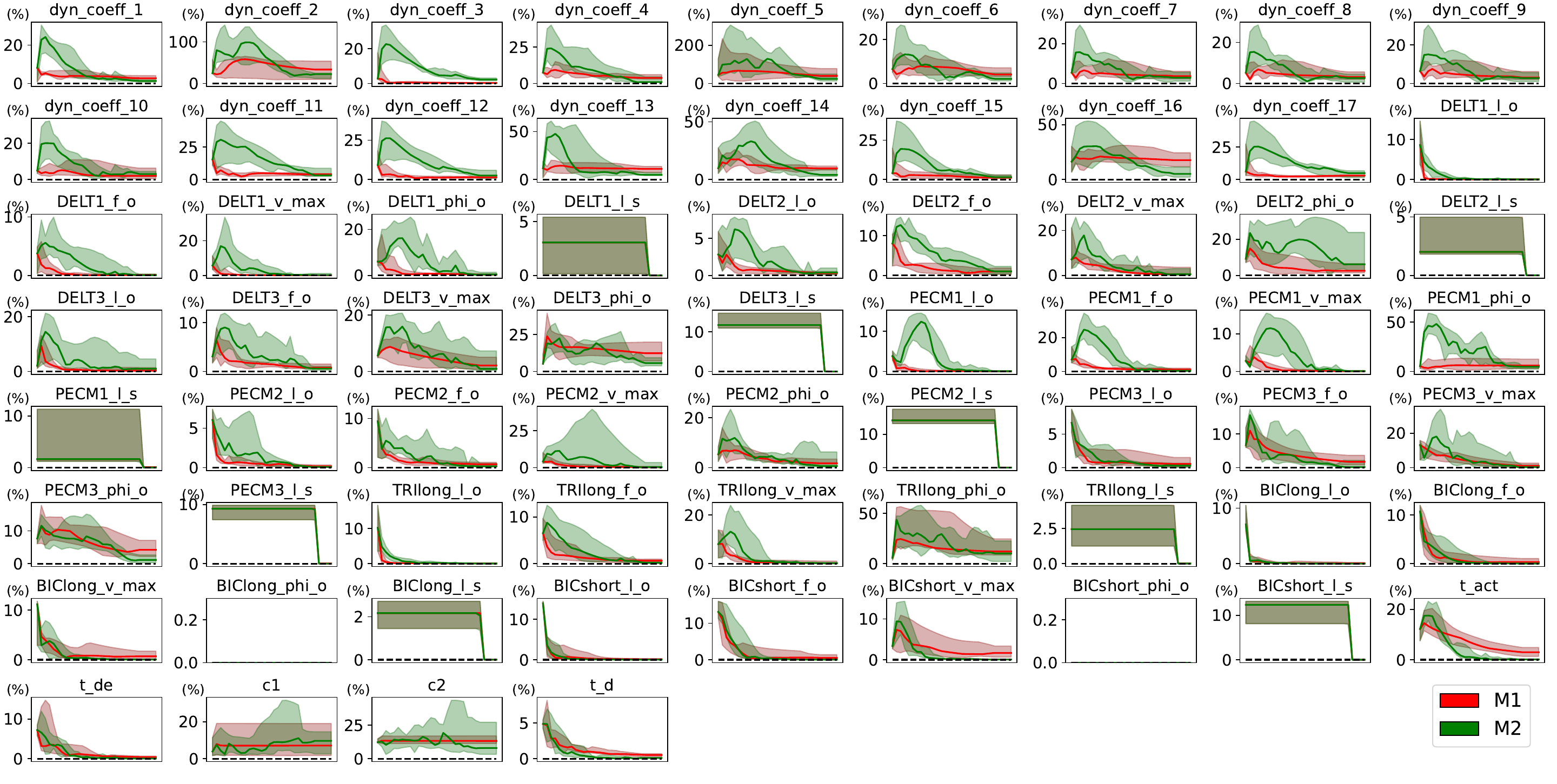} } 
	\vspace{-0.2cm}
	\caption{Percentage errors of the parameter estimates over 12k epochs for both methods M1 (red, without deep muscle EMGs) and M2 (green, with deep muscle EMGs). The solid line and shaded area represent the median, upper and lower quartiles of the percentage error.}
	\label{fig:parameters evolution}
	\vspace{-0.2cm}      
\end{figure*}

For each of the two arm movements, five sets of initial guesses for the learnable parameters were sampled from the normal distribution and used for each of the two methods. 
\begin{equation}
    \pi_{0} \sim N(\pi_{gt}, 0.1\pi_{gt}),
    \label{eq:initvaluesample}
\end{equation}
Each of the total 20 simulation runs was conducted for 10K epochs, followed by an additional 2K epochs to optimize the tendon slack length based on the other optimized parameters. All the simulations were completed on a workstation with an Intel Core I9-14900KS CPU and 64GB of memory. The computation time per epoch was about $0.77s$, yielding a total runtime of about $2.5$ hours for each run. The parameter estimation results of the comparative methods (M1 and M2) in terms of the three criteria are shown in Fig. \ref{fig:res:C1C2} and Fig. \ref{fig:parameters evolution}.

In Fig. \ref{fig:res:C1C2}, it can be observed that our proposed method M1 converged faster than M2 in terms of $Cr_1$. However, at the 10K epoch mark, the methods performed similarly in terms of all the criteria. With the final $Cr_2$, the performance of M1 was slightly worse than M2, while M1 achieved at least comparable results at the end of the optimization with $Cr_2^m$. It seems to imply that, with more learnable parameters in M2, the optimization problem was harder to solve but had the potential to find a better solution. However, the estimates of all the identifiable parameters of superficial muscles were already as good as M2. This is important as the bone parameter uncertainty has less impact on the dynamic uncertainty than the kinematic parameter uncertainty, shown in \cite{muller2017uncertainty}, but the muscle parameters greatly influence force-generating behavior of the musculotendon system \cite{winby2008evaluation}. 

There exists redundancy in the bone dynamic parameters, i.e., an infinite number of parameter solutions can achieve the same dynamic behavior of bones. However, a set of dynamic coefficients, as linear combinations of the original dynamic parameters, can be uniquely identified \cite{gaz2019dynamic}. These coefficients are a minimum set that can determine the dynamic behavior of bones. Therefore, the original 30 bone dynamic parameters were converted to a set of 17 dynamic coefficients using the approach in \cite{atkeson1986estimation}, and the convergence of these coefficients to their ground truth values can be seen in Fig. \ref{fig:parameters evolution}. It is clear that most parameters could be estimated decently well by both methods, and in most cases, parameters converged actually faster and more robustly with M1 compared to M2. However, some of the parameters, e.g. $c_1,c_2,\phi_o$, seemed to converge more slowly than the others for both methods. It is worth noting that \textit{BIClong} and \textit{BICshort} are not pennated muscles, therefore, their $\phi_o$ could not be set to a value other than their ground truth value of zero, as shown in Fig. \ref{fig:parameters evolution}.

\vspace{-0.1cm}
\section{Implications}

This simulation study is an important step for the practical use of the proposed parameter identification method with differentiable optimization, i.e., we considered the inaccessibility of the deep muscle activation and improved the realism of the musculoskeletal model, which simulates the real situation as much as possible. The anatomical features of the human arm (that is, moment arms of the deep muscles) were utilized to estimate the shoulder muscle/tendon parameters without deep muscle EMG reconstruction. It is valuable to accurately and invasively estimate the parameters of the shoulder-related muscles and tendons, as the shoulder typically needs to bear a large joint torque and is more prone to problems compared to other arm joints in physically demanding tasks. This method can enable, for instance, the convenient assessment of shoulder muscle strength and tendon stress-strain curve for many healthcare-related applications.

\vspace{-0.1cm}
\section{Conclusions and future work}

In this work, we proposed to simultaneously identify the bone and muscle parameters of a subject-specific human arm musculoskeletal model without reconstruction of the deep muscle EMGs. This was achieved by treating the deep muscle behavior as a black box, and only using the best possible least-squares solution of the deep muscle forces to calculate the loss gradient with respect to the model parameters for optimizing the parameters within a framework of differentiable optimization. Satisfying results were observed in comparative simulations. Although a good initial guess of the model parameters can be obtained with appropriate scaling methods \cite{winby2008evaluation, modenese2016estimation} based on a template model, more simulation tests with larger initial parameter errors and real experiments will be carried out in the future.

\bibliographystyle{IEEEtran}	
\bibliography{IEEEabrv,references}

\end{document}